\documentclass{article}
\usepackage[final,nonatbib]{neurips_2024}
\usepackage[utf8]{inputenc} 
\usepackage[T1]{fontenc}    
\usepackage{hyperref}       
\usepackage{url}            
\usepackage{booktabs}       
\usepackage{nicefrac}       
\usepackage{microtype}      
\usepackage{xcolor}         
\usepackage{listings}%
\usepackage{colortbl}
\usepackage{makecell}
\usepackage{subfigure}
\usepackage{graphicx}%
\usepackage{multirow}%
\usepackage{amsmath,amssymb,amsfonts}%
\usepackage{amsthm}%
\usepackage{mathrsfs}%
\usepackage{textcomp}%
\usepackage{algorithm}%
\usepackage{algorithmicx}%
\usepackage{algpseudocode}%
\newcommand{\ie}{i.e.,\ }

\newcommand{\aka}{\emph{a.k.a.},\ }
\newcommand{\et}{\emph{et al.}\ }

\title{Automated Design and Optimization of Distributed Filter Circuits using Reinforcement Learning}

\author{
Peng Gao$^{1}$, Tao Yu$^{1}$, Fei Wang$^{2}$, Ru-Yue Yuan\\
$^1$Qufu Normal University \quad
$^2$Harbin Institute of Technology Shenzhen
}

\begin{document}

\maketitle

\begin{abstract}
Designing distributed filter circuits (DFCs) is complex and time-consuming, involving setting and optimizing multiple hyperparameters. Traditional optimization methods, such as using the commercial finite element solver HFSS (High-Frequency Structure Simulator) to enumerate all parameter combinations with fixed steps and then simulate each combination, are not only time-consuming and labor-intensive but also rely heavily on the expertise and experience of electronics engineers, making it difficult to adapt to rapidly changing design requirements. Additionally, these commercial tools struggle with precise adjustments when parameters are sensitive to numerical changes, resulting in limited optimization effectiveness. This study proposes a novel end-to-end automated method for DFC design. The proposed method harnesses reinforcement learning (RL) algorithms, eliminating the dependence on the design experience of engineers. Thus, it significantly reduces the subjectivity and constraints associated with circuit design. The experimental findings demonstrate clear improvements in design efficiency and quality when comparing the proposed method with traditional engineer-driven methods. Furthermore, the proposed method achieves superior performance when designing complex or rapidly evolving DFCs, highlighting the substantial potential of RL in circuit design automation. In particular, compared to the existing DFC automation design method CircuitGNN, our method achieves an average performance improvement of 8.72\%. Additionally, the execution efficiency of our method is 2000 times higher than CircuitGNN on the CPU and 241 times higher on the GPU.
\end{abstract}

\section{Introduction}\label{sec:1}

Distributed filter circuits (DFCs) are designed by combining components such as resonators. Numerous domains, ranging from complex cellular networks to expansive satellite communication systems, rely heavily on these circuits \cite{r1,r2,r3}. In particular, in high-frequency scenarios, electronic components often exhibit impedance characteristics. An efficient DFC plays a crucial role in facilitating rapid signal transmission and clear signal reception. In contrast to a low-frequency analog circuit design that treats each component as an independent unit with a specific functionality, a DFC design involves interdependent components. The geometric parameters and spatial relationships of each component in DFCs can significantly impact the functionalities of other components. Moreover, the geometric parameters of one component can be influenced by other components \cite{r5,r6}. This interdependence among components can lead to unexpected resonances and scattering phenomena in DFCs.

\begin{figure}[t]
\begin{center}
    \subfigure[Six-resonator bandpass filter]
    {\label{fig:1a}\includegraphics[width=0.3\linewidth]{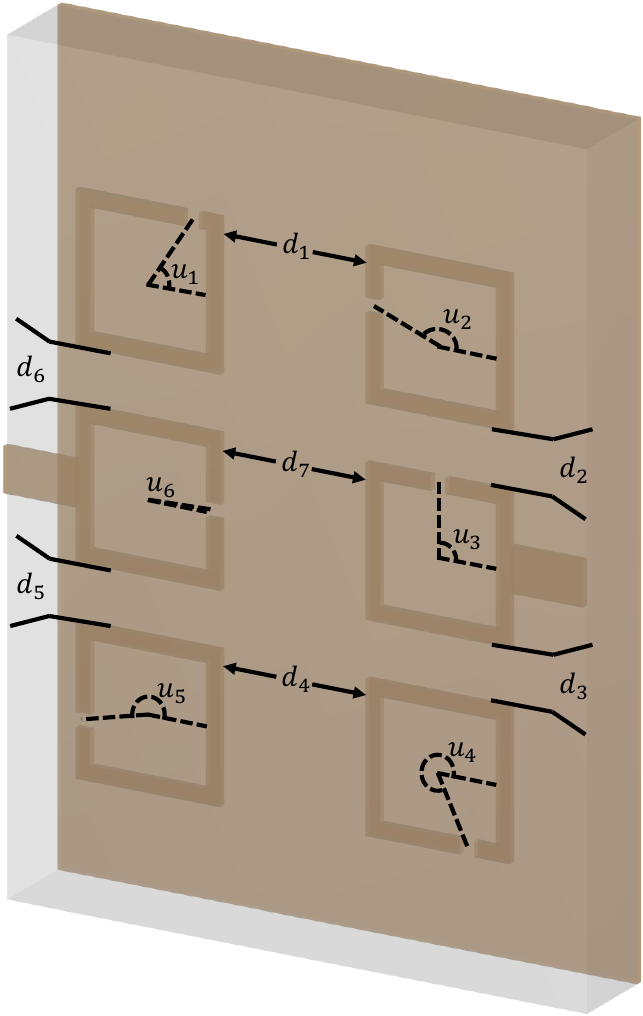}}
    \hspace{2em}
    \subfigure[Four-resonator bandpass filter]
    {\label{fig:1b}\includegraphics[width=0.4\linewidth]{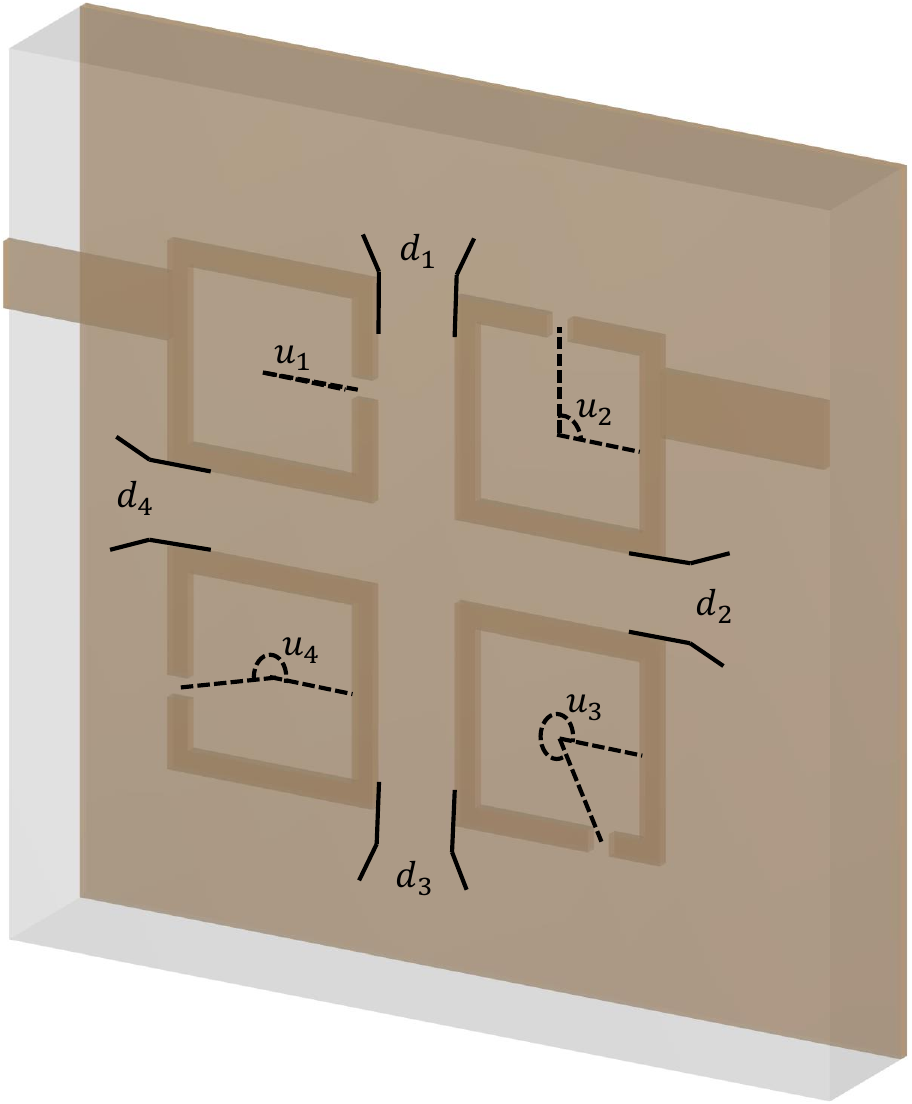}}
\end{center}
\caption{Examples of DFC templates based on square resonators. In the figure, $d$ indicates the relative distance between two resonators, and $u$ represents the counterclockwise angle from the horizontal line to the line connecting the opening position and the center point.}
\label{fig:1}
\end{figure}

A typical DFC comprises parameterized resonators, as depicted in Fig.\ref{fig:1}. In DFC design, squares \cite{r4} or circular resonators \cite{r5} typically serve as the fundamental building blocks. These resonators can selectively amplify or suppress specific signal frequencies based on the design requirements. Resonance occurs when the frequency of the input signal matches the resonant frequency of the resonator. Electronic engineers manually design DFCs, which is an intricate process that often involves the fulfillment of multifaceted requirements. An electronic engineer proposes the initial design scheme by considering the expected frequency passband of the filter circuit. Subsequently, the engineer generally selects a filter circuit template composed of resonators and iteratively adjusts and optimizes the geometric parameters of these resonators within the template. Moreover, engineers frequently conduct electromagnetic simulations via commercial software tools to predict signal-resonator interactions and those among the resonators themselves. However, this design process is time-consuming and resource intensive, and is predominantly dependent on the expertise and experience of electronics engineers.

With the continuous evolution of artificial intelligence (AI), reinforcement learning (RL) has emerged as an effective tool for solving decision-making problems. RL can adaptively adjust optimization strategies through continuous environmental interaction, gradually improving design quality. RL agents can find optimal solutions in complex hyperparameter spaces without exhaustively simulating all possible combinations. By learning and optimizing strategies, RL agents can effectively reduce the number of simulations and the time required, thereby improving optimization efficiency. \cite{r5} noted that different filter circuits can be designed by altering the number and geometric parameters of resonators. Leveraging this capability, we consider the design and optimization of DFCs as sequential decision-making tasks. The RL agent optimizes the performance of the DFCs by modifying the geometric parameters. However, it is challenging to design a DFC that fulfills the performance requirements. The primary obstacle is the highly complex nonlinear relationship between the geometric parameters of the circuit and the performance requirements. In most cases, the resonators must be positioned without overlap.

In this study, we developed a novel end-to-end automated method for DFC design to address the aforementioned problems. The main contributions of this study are outlined below:
\begin{itemize}
  \item Introduction of an RL-based DFC design and optimization (RLDFCDO) method that automatically adjusts the geometric parameters of individual resonators to optimize circuit performance.
  \item Implementation of the best reward initialization (BRI) method, facilitating the design of initial circuit layouts compliant with performance requirements yet non-optimal.
  \item Introduction of the state evaluator (SE) method to evaluate the performance of the circuits designed by RLDFCDO and BRI.
  \item Utilization of an invalid action penalty (IDP) mechanism that enables RLDFCDO to avoid invalid design schemes (overlapping the placement of resonators), thereby increasing the probability of RL agents learning superior circuit design schemes.
  \item Provision of an end-to-end DFC design and optimization solution. A filter circuit design that satisfies the stipulated requirements can be obtained by inputting the expected frequency passband.
\end{itemize}

The rest of the paper is organized as follows: we first introduce the related works in Section \ref{sec:2}, then show how to design and optimize DFCs via reinforcement learning in Section \ref{sec:3}, following conduct extensive experiments in Section \ref{sec:4} to evaluate the proposed method, and finally draw a conclusion in Section \ref{sec:5}.

\section{Related Works}\label{sec:2}

\subsection{Learning-Based Circuit Design and Optimization}\label{sec:2-1}

Historically, circuit design and optimization have presented challenges, stimulating innovative methodologies. \cite{r22} introduced geometric programming to optimize phase-locked loop circuits. \cite{r23} proposed an analog circuit optimization approach based on a hybrid genetic algorithm (GA), showcasing the potential of GAs in this domain. With the evolution of machine learning technology, the direction of research on circuit design and optimization methods has gradually shifted. \cite{r27} combined sparse regression and polynomial optimization to synthesize analog circuits efficiently. \cite{r29} and \cite{r30} further explored the potential of GAs in circuit optimization. McConaghy \et focused on trustworthy genetic programming-based synthesis using hierarchical domain-specific building blocks. Meanwhile, Lourencco \et introduced GENOM-POF, a multi-objective evolutionary synthesis approach with corner validation. Subsequently, \cite{r24,r25,r26} presented a series of Bayesian optimization-based analog circuit design methods, highlighting their efficacy in handling uncertainty and optimizing complex design spaces. \cite{r28} investigated end-to-end learning for distributed circuit design and graph neural networks (GNNs) \cite{r6}, demonstrating the potential of deep learning techniques for abstracting complex state-space information and optimizing circuit layouts. Furthermore, \cite{r31} and \cite{r32,r33} employed complex neural networks and parameter modeling of transfer functions. They emphasized the significance of accurate modeling in ensuring the optimized design and performance of microwave components. Recent studies have highlighted the immense potential of deep learning in automated circuit design. For instance, some researchers have employed deep neural networks (DNNs) based on Thompson sampling for efficient multi-objective optimization \cite{r34}. Others have proposed an RL-based quantum circuit optimizer \cite{r35}, employing gradient-based supervised learning algorithms to train a design agent to satisfy the threshold specifications of analog circuits \cite{r36}.

\subsection{RL for Sequential Decision-Making Tasks}\label{sec:2-2}

Over the past decade, significant advancements have made RL algorithms adept at effectively addressing challenging sequential decision-making problems. Early RL research focused primarily on value-based methods. The value function serves as a core concept for estimating the expected return of an agent to a given state or after taking a specific action. Techniques such as Q-learning \cite{r7} represent the value function in a tabular manner, making action decisions by maximizing this function. Although table-based methods are effective in small, discrete environments, they struggle to scale up to more extensive or continuous state spaces. Linear function approximation methods such as tile coding enable RL to handle large state spaces \cite{r8,r9}. However, tile coding has some fatal flaws, primarily in its fixed and non-learning-capable tiles and the difficulty involved in representing complex state features through linear combinations of tiles. These drawbacks of tile coding and other linear approximation methods have motivated scholars to explore nonlinear function approximations, such as neural networks \cite{r10}. Nevertheless, early shallow neural networks could not easily achieve the representational power and stability attained in traditional RL. A major turning point in this field was the introduction of deep RL (DRL), which automatically extracts useful state features by utilizing DNNs. Neural networks enable the complex abstraction modeling of complex state spaces. End-to-end training enables neural networks to adapt to RL problems. DRL has paved the way for new possibilities in addressing a wide array of sequential decision-making challenges. It primarily employs DNNs for functional approximation. Deep Q-networks (DQNs) combine vital methods such as Q-learning, DNNs, and experience replay \cite{r11}, enabling RL agents to master complex visual environments and achieve human-level performance. Double DQN \cite{r12} and dueling architectures \cite{r13} have further improved the performance of DRL. DRL has been applied to various sequential decision-making tasks, including robotics \cite{r14}, games \cite{r15}, air combats \cite{r49}, architectural design \cite{r50}, and recommendation systems \cite{r16}.
Another crucial approach is the actor-critic method \cite{r17}, which requires separate maintenance of the policy and value functions. The asynchronous advantage actor-critic approach \cite{r18} achieved state-of-the-art results by parallelizing experience collection. Policy gradient techniques, such as trust region policy optimization (TRPO) \cite{r19} and proximal policy optimization (PPO) \cite{r20}, directly optimize neural network policies while ensuring monotonic yet stable training improvement. \cite{r37} introduced RL in large-scale photonic recurrent neural networks, demonstrating outstanding predictive performance in the Mackey–Glass chaotic time series. \cite{r38} proposed Swift, an autonomous drone racing system that utilizes real-world data and fine-tunes the model using the PPO algorithm. \cite{r21} explored the decomposition of problems into higher-level abstractions by using hierarchical RL.
Recently, RL has displayed extensive potential and made significant strides across various fields. \cite{r39} investigated the metacognitive processes of motor learning using RL, revealing how humans optimize learning strategies by adjusting learning and retention rates based on feedback. \cite{r40} applied RL to skin cancer diagnosis, considerably improving the diagnostic sensitivity and accuracy by integrating medical expert preferences with non-uniform reward and penalty mechanisms. \cite{r41} employed a stochastic subsampling RL method to optimize long-term career path recommendation systems. Regarding robotics control, \cite{r42} successfully transferred policies trained in simulation environments to the real world. \cite{r43} effectively controlled complex laser systems using deep RL, offering new technological pathways for stable pulse generation. These studies showcase the extensive applicability of RL across multiple domains and highlight its remarkable achievements through innovative approaches.

\section{Proposed Method}\label{sec:3}

\begin{figure}[t!]
    \begin{center}
        \includegraphics[width=\linewidth]{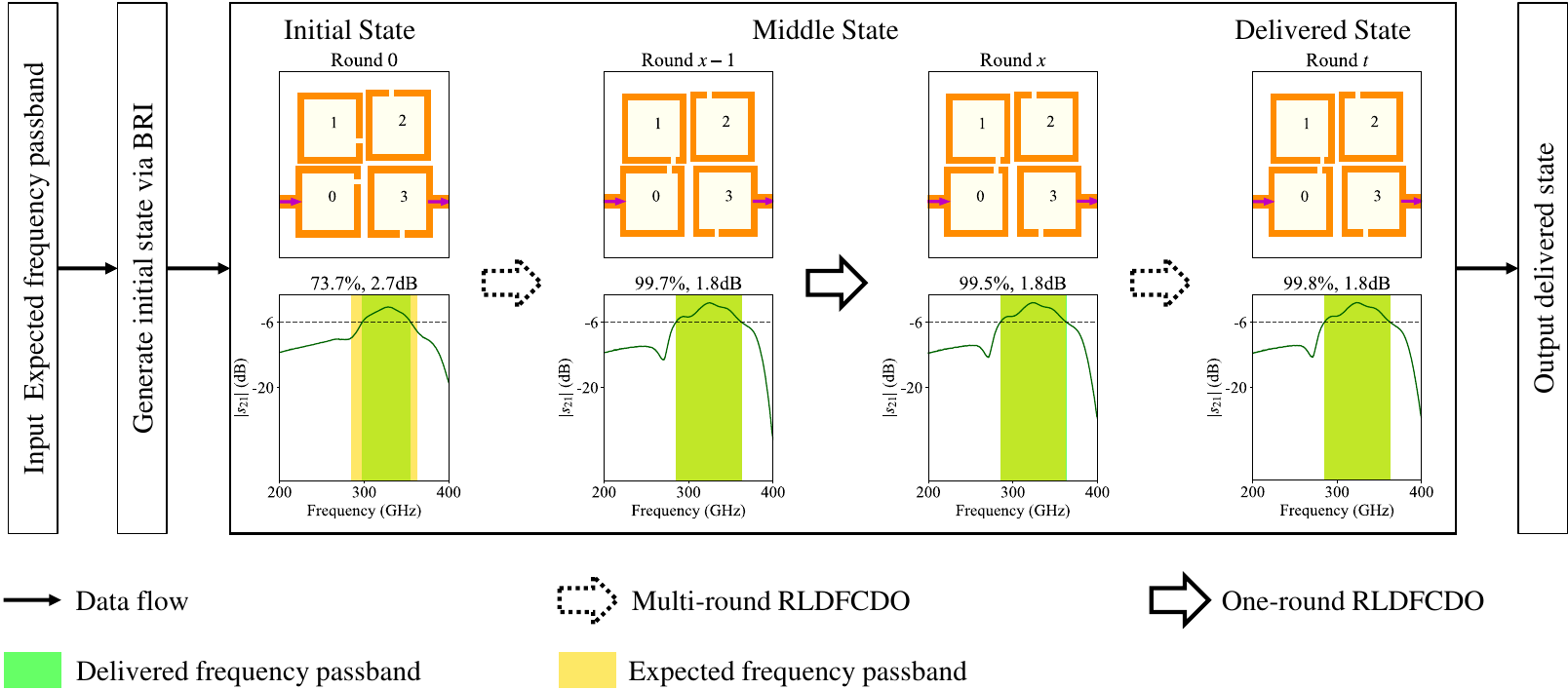}
    \end{center}
    \caption{Flowchart of end-to-end design and optimization of DFCs.}
    \label{fig:2}
\end{figure}

The geometric parameters of the resonators often determine the filter circuit performance. We aim to automate the optimization of these geometric parameters based on the expected frequency passband. The optimization process can be modeled as a Markov decision process, wherein an RL agent interacts with the environment, performs specific actions impacting it, and receives rewards based on environmental feedback. Through multiple iterations and interactions with the environment, the agent learns an optimal policy that maximizes the long-term cumulative rewards. In DFC design and optimization problems, the environment, state, actions, and reward correspond to the circuit performance simulation, current geometric parameters of all resonators, changes in the geometric parameters, and circuit performance, respectively. The flowchart shown in Fig.\ref{fig:2} illustrates the basic process of optimizing circuit parameters based on a given expected frequency passband. First, input the expected frequency band for the DFC. For instance, to design a filter circuit with an expected frequency passband of (285, 363) GHz, use the BRI method to generate an initial circuit layout state that roughly meets the requirements, as shown in Round 0. Then, use the RLDFCDO method over multiple rounds to modify the current circuit's geometric parameters to optimize its performance. In each round, the initial state is the best-performing layout from the previous round. The optimization continues until the specified number of rounds is completed or the circuit performance meets the expected requirements. Finally, the optimized circuit layout state is delivered, as shown in Round \textit{t}.

We employed the PPO algorithm \cite{r20} to train the agents. Compared to TRPO \cite{r19}, PPO offers several advantages, such as simpler implementation, higher and more stable sampling efficiency, and reduced sensitivity to hyperparameter selection. In this study, the optimization problem of the resonator geometric parameters was abstracted as a controllable RL process, establishing a novel optimization workflow for designing DFCs. The proposed method employs computer-based large-scale simulation for rapid iterative parameter optimization. As RL does not rely on the domain expertise and design experience of engineers, it can explore potential design solutions more efficiently than manual methods. Moreover, the approach only requires the specifications of the expected frequency passband. An RL method capable of automatically designing and optimizing DFCs based on the demands and constraints was implemented by constructing the RL agent (\aka policy network), the state space, the action space, and the reward scheme.
For convenience, we describe a design method that utilizes square resonators as the fundamental components of a DFC.

\subsection{State Space}\label{sec:3-1}

In RL methods, the state space directly affects the learning efficacy and is crucial in problem modeling. We designed a two-dimensional (2D) continuous state space to comprehensively reflect the geometric and spatial information of the circuit. A continuous state space effectively represents acceptable changes in the circuit parameters. In such a state space, different states correspond to minor variations in the circuit, enabling the agent to explore the design space intricately to identify the optimal parameter settings. Moreover, employing a continuous state enables leveraging of the advantages of RL methods for exploring optimal solutions within continuous spaces. Furthermore, continuous states naturally handle novel states that are not encountered during training, featuring strong generalization capabilities. The state space employs real-valued vectors to represent the geometric parameters of each resonator in the circuit.
Each resonator is characterized by an eigenvector composed of nine parameters: the horizontal position ($x$), vertical position ($y$), side length ($l$) of the square resonator cavity, wall thickness ($w$), horizontal position of the opening (gap$_x$), vertical position of the opening (gap$_y$), gap height (gap$_{h}$), gap width (gap$_{w}$), and the counterclockwise angle ($u$) from the horizontal line to the line connecting the opening position and the center point. These parameters reflect critical information regarding the spatial distribution and geometric shape of the resonator, which directly determines the circuit topology. As the resonator size parameters often have a significant impact on the frequency response of the circuit, the state space design should include as many relevant parameters as possible.
In a specific implementation, we represent the state space as a matrix, where each row vector corresponds to the feature representation of a resonator. The state matrix is fed into the policy network, where a fully connected NN with two hidden layers is employed to extract and transform the state inputs, generating a high-dimensional abstract representation of the circuit. Compared with a simple state representation, our 2D state space enables more effective learning, leading to more optimal circuit design schemes.

\subsection{Action Space}\label{sec:3-2}

Following state space creation, we designed a corresponding action space. The action space contains a set of actions that can be applied to the current circuit state to search for an optimal design by changing the circuit parameters. We employed a representation of a discrete action space for the design and optimization problem addressed in this study. In complex environments with numerous possible actions, using a discrete action space simplifies the learning process and enhances efficiency \cite{r44,r45}. This choice was primarily based on the following factors. First, when considering the multitude of optimizable parameters for the resonators, using a continuous action space burdens the learning process of the policy network, making it challenging to implement effective training. Discrete actions provide clear guidance for the search direction and reduce the learning difficulty. Second, discrete actions are advantageous for implementing a penalty mechanism for invalid actions and avoiding erroneous designs that are difficult to judge with continuous actions. Moreover, compared with the random fluctuations of continuous actions, discrete actions explicitly represent the range and precision of changes, reducing ineffective randomness and favoring stable learning. Finally, discrete actions can express a richer set of actions through combinations, enhancing the expressive ability.

Based on these considerations, we defined various combinations of discrete actions to optimize the geometric parameters of the resonator. These actions were applied to the specified resonators and could be freely combined to alter the circuit topology. Moreover, we explicitly outlined the effects of different geometric parameters on the circuit performance and appropriately set the variation amplitude for each action. For parameters that have a significant impact, we set small variation ranges. In contrast,  we allowed larger variation ranges for parameters less sensitive to minor adjustments. Moreover, we defined different combination actions to simultaneously adjust the different resonator parameters. In the later stages of training, the probability of combination actions is gradually increased for detailed optimization. Since the resonators cannot overlap, the feasibility of the action space is restricted. Therefore, we defined different penalty intensities for invalid actions based on the action range. However, a few invalid designs were allowed to maintain the completeness of the state space. We could rapidly explore adequate state space and develop a strategy for generating high-quality circuits by precisely setting the penalty intensity. In addition, we devised a specific operation wherein invalid actions were randomly executed with a certain probability, and hidden samples were introduced to enhance the generalization capability of the model. Additionally, we expanded the action space by incorporating combined operations that led to more varied changes in circuit topologies.

\subsection{Reward Design}\label{sec:3-3}

In RL methods, the reward mechanism is the core driver that guides the training of the model. We specifically crafted a multi-objective, dynamically adjustable reward function to induce the RL agent to autonomously optimize the circuit design parameters. The reward function comprises three primary components.

\begin{enumerate}
\item The first component is the frequency passband overlap. $\text{IOU}=\frac{\text{Pass}_t\cap\text{Pass}_d}{\text{Pass}_t\cup\text{Pass}_d}$ is calculated based on the passband $\text{Pass}_d$ of the delivered circuit and the expected passband $\text{Pass}_t$. This indicator mirrors the quality of the frequency passband coverage of the delivered circuit. A higher IOU signifies a greater overlap between the expected and delivered on-state frequencies.
\item The second component is the insertion loss, which measures the signal strength reduction. This loss quantifies the signal transmission efficiency from the input to the output and is generally measured in decibels (dB). In designing radio frequency (RF) and microwave components, a low insertion loss implies that the signal maintains high strength during transmission. We utilized the inverse of the maximum value of the circuit frequency response function as the insertion loss \cite{r6}. Design schemes with low losses were also rewarded.
\item The third component is the central frequency deviation, which reflects the concentration level of the frequency passband response. A smaller deviation indicates that the frequency passbands have a more concentrated alignment with the target center. This indicator, distinct from the IOU, guides the frequency passband to move closer to the central frequency.
\end{enumerate}

These three indicators can be combined in a linearly weighted manner to form a comprehensive reward function:
\begin{equation}\label{eq:1}
  R_t=\text{IOU}+\alpha\times(6+\max s_{21})-\beta\times|\text{DC}-\text{TC}|
\end{equation}
where $s_{21}$ denotes the frequency loss, which represents the gain or attenuation between different frequency signals and their responses; \text{DC} indicates the central frequency of the current circuit; \text{TC} refers to the expected central frequency. Generally, the insertion loss of the filter circuit is less than $6$ db to have a normal working frequency band. We stipulate that the insertion loss of the filter circuit should be less than $6$ dB for a normal working frequency passband. Since the loss $s_{21}$ is always less than $0$, the value of $(6+\max s_{21})$ is less than $6$, and the range of IOU is $[0, 100]$. To highlight the impact of insertion loss on rewards, the insertion loss is multiplied by the hyperparameter $\alpha$, and the central frequency deviation is multiplied by the hyperparameter $\beta$. Thus, the frequency passband IOU, insertion loss, and frequency concentration could be comprehensively assessed while conducting parameter optimization. Moreover, we dynamically adjusted the weights of the three components, initially optimizing the insertion loss to acquire circuit design schemes that may not satisfy the requirements but can function normally, then directing the working central frequency toward the desired one, and finally reinforcing the IOU to obtain feasible solutions. Additional penalties are imposed on invalid designs. Furthermore, the reward is gradually lowered during training to prevent overfitting. Following sufficient training, the RL agent learns strategies for configuring circuits that satisfy various criteria.
Evaluation of the central frequency deviation is an innovative aspect of the reward function. In addition to considering the degree of frequency passband overlap, it is essential to match the central frequency. We aim to encourage the agent to acquire the ability to centrally regulate the frequency passband distribution by introducing a central frequency deviation. The agent can modify the concentration of the frequency passband distribution by adjusting the resonator parameters when the deviation in the central frequency is adjusted. Thus, when faced with newly set desired central frequencies, the agent can quickly configure the corresponding circuit parameters, aligning the response with the desired central frequency without retraining from scratch.

\subsection{Feedforward Network Design}\label{sec:3-a}

\begin{figure}[t!]
    \begin{center}
        \includegraphics[width=\linewidth]{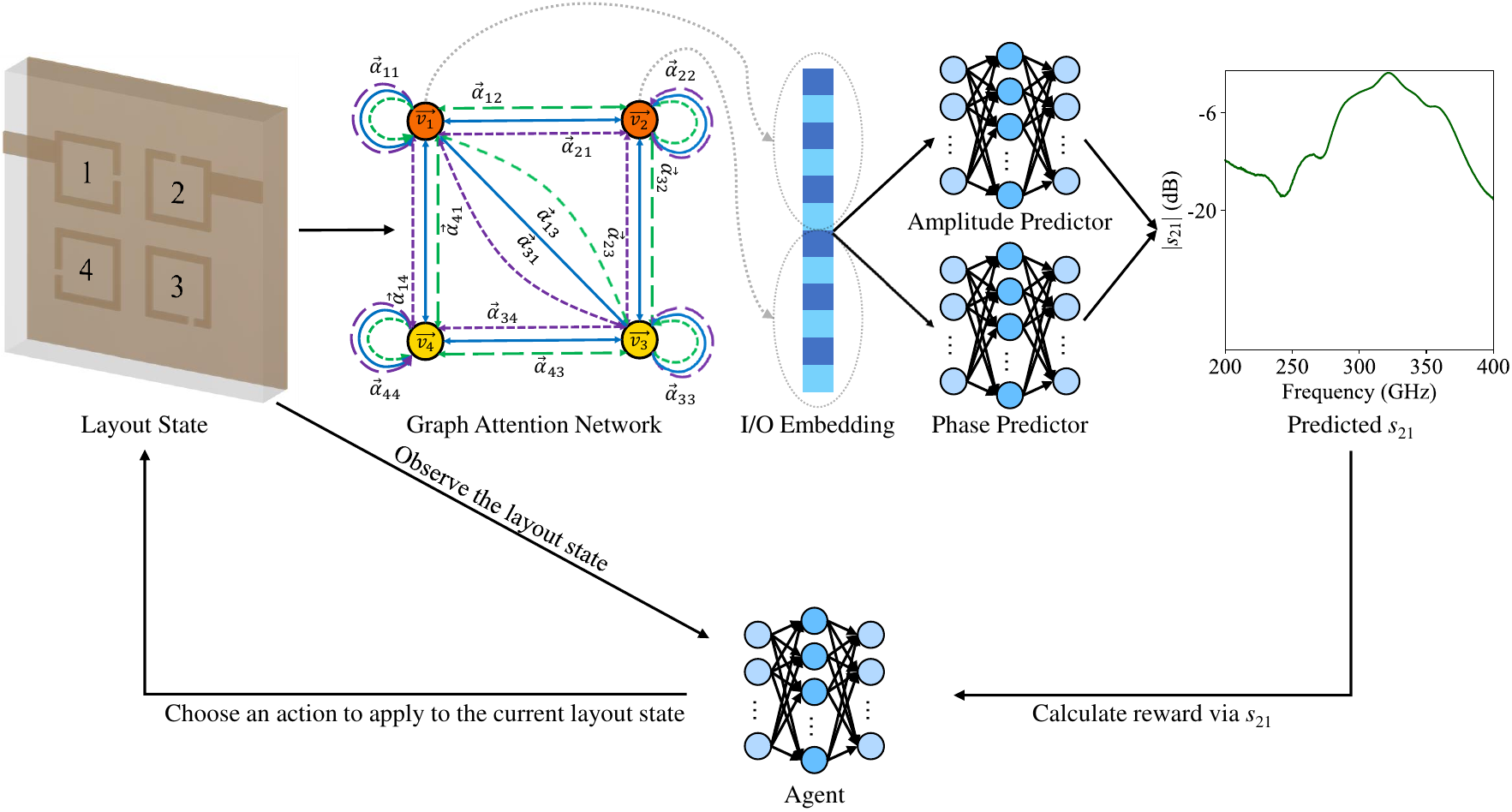}
    \end{center}
    \caption{Overall architecture of our proposed method. The black solid lines represent the data flow, and the three different colored lines (blue, purple, and green) in the GAT represent three independent attention computations, \ie three-head attention. The gray dashed line indicates the concatenation of the embeddings learned from the input and output resonators of the DFC, which then serves as the input for the next stage of amplitude and phase prediction.}
    \label{fig:9}
\end{figure}

The feedforward network aims to quickly evaluate the designed DFC's performance. It takes the geometric parameters of the DFC as input and outputs the $s_{21}$ indicator of the circuit. The upper part of Fig.\ref{fig:9} illustrates the detailed architecture of the forward network. There are four main steps from input to output:

\begin{enumerate}
    \item \textbf{From circuit layout to graph representation:} We map the geometric layout of the DFC to a graph representation, where each node in the graph represents a resonator, and edges between nodes represent interactions (\ie electromagnetic coupling) between pairs of resonators. A circuit with $N$ square resonators has $N$ initial parameter vectors. Each square resonator is represented by a vector of nine parameters [$x$, $y$, $l$, $w$, gap$_x$, gap$_y$, gap$_{h}$, gap$_{w}$, $u$], as detailed in Section \ref{sec:3-1}. Based on this initial input, we preprocess the geometric layout of the DFC into a graph representation with nodes and edges. Notably, not all node pairs (\ie resonator pairs) in the graph have edges. Since electromagnetic coupling decreases with distance, we set a threshold similar to \cite{r5}, beyond which there is no edge between two nodes.
    \item \textbf{Embeddings extraction using graph attention network (GAT):} We use GAT \cite{r47} to extract embeddings from the nodes and edges in the graph. Specifically, we use a 4-layer GAT to process the input embeddings. In each layer, we first perform a linear projection on each node's embeddings, denoted as $h^\prime_i=Wh_i$, where $W$ is a learnable weight matrix. Next, we compute the attention coefficient $e_{i,j}=$LeakyReLU$(a^T[h^\prime_i||h^\prime_j])$ between the node $i$ and its neighboring node $j$ and normalize it using the Softmax function to obtain $a_{i,j}=\frac{\exp{e_{i,j}}}{\sum_{k\in N_i}\exp{e_{i,k}}}$, where $a$ is a shared attentional mechanism defined by \cite{r47}, $N_i$ denotes the set of all neighbors of node $i$. We then use these attention coefficients to compute a weighted sum of the neighboring nodes' embeddings, resulting in updated node embeddings $h^{\prime\prime}_i=\sigma(\sum_{j\in N_i}\alpha_{i,j}h^\prime_j)$. For each GAT layer, we use 3 attention heads, corresponding to the blue, purple, and green colors in Fig.\ref{fig:9}. Each head independently computes attention coefficients and updates embeddings, and the final embedding for node $i$ is the average of the attention heads' outputs.
    \item \textbf{Summarizing processed graph data:} We transform the embeddings of the graph into fixed-length global graph embeddings $G$ by concatenating the embeddings of two particular nodes. These two nodes represent the resonators connected to the input and output of the DFC.
    \item \textbf{Predicting the real and imaginary parts of $s_{21}$:} Using the global graph embeddings $G$, we predict the real and imaginary parts of the $s_{21}$ indicator with the amplitude and phase predictors, respectively.
\end{enumerate}

We use the traditional supervised learning paradigm to train our feedforward network. During the training process, we employ L1 loss. This encourages the network to learn sparse embedding representations, improving generalization ability and interpretability. Additionally, we incorporate cross-validation to evaluate the network performance and dynamically adjust the learning rate while using an early stopping strategy to prevent overfitting.

\subsection{Policy and Value Networks Design}\label{sec:3-4}

Policy and value networks are the core components of our proposed method, as shown in Fig.\ref{fig:9}. The policy network (\aka the RL agent) generates the probability distribution of the actions to be taken in a particular state. In the proposed approach, the policy network includes an input layer, two hidden layers, and an output layer. The input layer receives representations of the geometric circuit parameters. The number of nodes is equal to the dimensionality of the state representation. The first hidden layer includes $256$ fully connected nodes, each connected to all nodes of the input layer. It introduces nonlinearity through a rectified linear unit activation function to extract advanced-state features. Similarly, the second hidden layer contains $256$ fully connected nodes connected to all nodes of the first hidden layer. The number of nodes in the output layer corresponds to the action space. A \text{Softmax} activation function transforms the value of each node into its corresponding action probability. This multi-layer, fully connected network structure effectively integrates state feature learning and action probability output. This policy-network structure is widely adopted for RL. Furthermore, to enhance the generalization capability of the network, the first hidden layer includes a dropout operation that randomly masks certain nodes to prevent overfitting.
The value network evaluates the long-term return of each state-action combination and outputs the corresponding state-action value. The value network adopts a similar multi-layer, fully connected structure. It uses the geometric parameter representation of the circuit as the input and predicts the state-action value using two hidden layers.

\subsection{Environment Design}\label{sec:3-5}

\subsubsection{Environment Creation}\label{sec:3-5-1}

In a designed environment, a pivotal step is to provide a circuit-parameter configuration scheme for the initial state of the environment. Accordingly, we devised two approaches to obtain high-quality initial states.

The first approach is the BRI, which generates a circuit topology that reasonably satisfies the performance requirements. Specifically, we utilized CircuitGenerator \cite{r6} to generate $2000$ circuit topologies randomly. The score for each structure was calculated as,
\begin{equation}\label{eq:2}
  R_t=\text{IOU}+\alpha\times(6+\max s_{21})
\end{equation}
and the structure with the highest score was selected as the initial state of the environment. The proposed method continuously optimizes the generated initial circuit topology, reducing memory consumption and improving efficiency.

Another approach involves manually specifying the initial state of an environment. This method enables the use of circuit topologies designed manually by electronic engineers based on their experience as the initial state of the environment. It also permits further optimization based on the previous round of optimization, as detailed in Section \ref{sec:3-6}.

\subsubsection{Environment Control}\label{sec:3-5-2}

In the designed circuit optimization environment, the action space is discrete and different sensitivity levels exist for the optimized parameters. The magnitude of the changes in these parameters causes varying degrees of state transitions owing to the various actions taken, leading to disparate results. Therefore, a robust state-control strategy must be developed. Effective state control ensures a smooth progression of the training process. The variation factors for all operational parameters are set to $0.1$.
The primary logic for state control is as follows:
\begin{enumerate}
    \item Initially, the environment maintains a triplet comprising the initial circuit state, initial circuit frequency response $s_{21}$, and initial circuit reward. Similarly, it maintains another triplet comprising the current circuit state, current circuit frequency response $s_{21}$, and current circuit reward. When the agent performs a new action, the horizontal and vertical positions of the resonators are less sensitive to numerical variations. In this study, the variation factor was set to $1$. The side length and thickness of the resonator cavity are relatively sensitive to numerical variations and were thus assigned a variation factor of $0.5$. The counterclockwise angle $u$ from the horizontal line to the line connecting the opening position and the center point was the most sensitive to numerical variations. Thus, its variation factor was set to $0.01$ and could be changed cyclically. We ensured that upon altering $u$, its value fell within the range of $[0, 1]$ by adjusting it through periodic increments or decrements.
    \item Subsequently, the environment verifies the validity of the new state. For parameter conflicts, the environment applies a negative reward penalty based on an IDP policy. The valid states are subjected to a performance evaluation. Step rewards are computed based on the current state reward and the reward before the change. Moreover, the environment manages the initiation and termination of training episodes. Once an episode begins, the environmental state is initialized. Specifically, during the first call of the initialization method, the BRI method generates a circuit topology as the initial state. Throughout the iterative optimization of the parameters by the agent, the environment assesses whether the optimization objective has been achieved. If achieved, it concludes the current episode and triggers the storage module to save important training data for this iteration, including the state sequence, rewards, and total rewards. The corresponding data serve algorithm optimization and analysis purposes.
\end{enumerate}

\subsubsection{Performance Evaluation}\label{sec:3-5-3}

We designed a performance evaluation scheme for the proposed method. Specifically, we predicted $s_{21}$ for the current circuit parameter configuration by utilizing the feedforward network designed in Section \ref{sec:3-a}. Subsequently, we calculated the passband IOU based on $s_{21}$ and considered the maximum value of $s_{21}$ as the insertion loss of the current scheme for circuit parameter optimization. The frequency at which $s_{21}$ reaches its maximum value becomes the working central frequency of the current scheme for circuit parameter optimization. We aimed for this evaluation method not only to optimize single-frequency passband design but also to address challenges posed by the optimization of multi-frequency passband design.
For a single-frequency passband, the IOU calculation is relatively simple and only involves intersecting and comparing the delivered frequency passband with the expected frequency passband. However, multi-frequency passband IOU calculations are more intricate, as interactions between the frequency passbands must be handled. Specifically, when optimizing multiple frequency passbands simultaneously, calculating the IOU separately for each passband and performing averaging is inadequate. One challenge is that the delivered frequency passband of the circuit may not entirely overlap with the expected frequency passband; instead, partial overlap can occur. In the delivered circuit, any frequency passband that does not overlap with the target circuit is considered ineffective. Simple IOU calculations may be erroneously included within the effective overlap range. Another challenge is managing the distance relationships between the frequency passbands. Long inter-band intervals offer no benefit to model learning, requiring the consideration of penalties or adjustments to the frequency passband weights. The frequency passband positions must also be appropriately managed. If multi-frequency passbands are included in the same region, then the problem of repeated calculations must be addressed. To address this problem, we propose a dynamic programming algorithm for calculating the multi-frequency passband IOU.

\begin{algorithm}
\caption{Calculate multi-frequency passband IOU}\label{alg:1}
\begin{algorithmic}[1]
\Require $X$: the list of effective frequency passbands for the currently designed circuit; $Y$: the list of frequency passbands for the expected circuit; $L_X$: the length of $X$; $L_Y$: the length of $Y$
\Ensure Multi-frequency passband IOU
\State Initialize a zero-array IOUs of size $(L_X+1) \times (L_Y+1)$
\For{$i = 1,\dots,L_X$}
    \For{$j = 1,\dots,L_Y$}
        \State Compute the single-frequency passband intersection $\text{I}=\text{Pass}_{i}\cap \text{Pass}_{j}$
        \State Compute the single-frequency passband union $\text{U}=\text{Pass}_{i}\cup \text{Pass}_{j}$
        \If{$\text{U} > 0$}
            \State $\text{IOU}=\text{I}/\text{U}$
            \State $\text{IOUs}[i,j]=\max(\text{IOU}, \text{IOUs}[i-1,j], \text{IOUs}[i,j-1])$
        \Else
            \State $\text{IOUs}[i,j]=\max(\text{IOUs}[i-1,j], \text{IOUs}[i,j-1])$
        \EndIf
    \EndFor
\EndFor
\State \Return $\text{IOUs}[L_X,L_Y]$
\end{algorithmic}
\end{algorithm}

\subsection{End-to-End DFC Design Automation}\label{sec:3-6}

To fully automate the end-to-end generation of circuit schemes from design objectives, we developed an efficient and stable RL-based method that considers the effectiveness, flexibility, and diversity within a discrete action space. The system was established by integrating comprehensive state representations and reward mechanisms and operated solely based on the input of the expected frequency passband. Our approach entails two key stages: rapid, stable, and high-quality initialization, followed by meticulous step-by-step optimization.
In the initial stage, the BRI rapidly generates multiple circuit topologies. The optimal topology is then selected using a performance evaluation function, which serves as the starting point for the subsequent optimization. Our end-to-end automated design method does not require a highly specific initial state for the circuit. It can iteratively progress from a deplorable initial state to achieve an excellent circuit design.
The second stage performs step-by-step optimization using deep RL. We trained an RL agent, RLDFCDO, that interacts with a circuit simulation environment akin to an experienced electronic engineer in the real world. It adjusts the parameters incrementally based on performance feedback, gradually approaching the target frequency response. We encapsulated this agent as a callable module that could be seamlessly embedded into the automation process, enabling fine-grained automatic optimization.
Moreover, we designed a comprehensive control logic that autonomously generated the initial states from the initialization model BRI and imported them into RLDFCDO for iterative optimization. Additionally, we developed a performance indicator monitoring and evaluation mechanism to terminate iterations actively and output the final results. Combining efficient prior-driven initialization and stable RL optimization resulted in an end-to-end closed-loop automation process.

\section{Experments and Results}\label{sec:4}

\subsection{Benchmarks and Settings}\label{sec:4-1}

\begin{table}[t!]
\centering
\caption{Details of the open-loop square resonator dataset.}
\label{tab:a}
\begin{tabular}{lllll}
\toprule
\#Resonator & 4 & 5 & 3 & 6 \\
\midrule
\#Sample  & 150,000 & 225,000 & 4,000 & 5,400  \\
\bottomrule
\end{tabular}%
\end{table}

We use the open-loop square resonators dataset provided by CircuitGNN to train our feedforward network. This dataset contains numerous DFC samples with geometric parameters and corresponding $s_{21}$ indicators. Each sample set includes various topologies, with the dataset detailed in Table \ref{tab:a}. We use 80\% of the data for DFCs composed of 4 and 5 resonators from the dataset for training and the remaining data for testing. The data for DFCs consisting of 3 and 6 resonators are entirely reserved for testing. During training, we used the Adam optimizer with a batch size of 128. The feedforward network is trained for a total of 500 epochs. The initial learning rate is set to $10^{-4}$ and decays by 0.5 every 200 epochs. We implemented the PPO algorithm using PyTorch \cite{r46} and constructed a circuit optimization environment named \emph{FilterEnv} based on our designed feedforward network. The circuit performance was evaluated based on the weighted sum of the IOU and insertion loss (Eq.\ref{eq:2}). All the validation experiments used a consistent learning rate of $7\times10^{-4}$ and were conducted on an NVIDIA GeForce RTX 3090 GPU with 24 GB VRAM. The RL agent randomly selected 200,000 samples during pre-training and employed our feedforward network to predict their corresponding $s_{21}$ indicators. The hyperparameters $\alpha$ and $\beta$ in Eqs.\ref{eq:2} and \ref{eq:3} were set to 10 and 1, respectively, and the number of random seeds was set to 8 before each experiment. Importantly, all the experimental results were reproducible.

\subsection{Ablation Studies}\label{sec:4-2}

We randomly selected four expected single-frequency passband targets: (240, 250), (250, 270), (270, 300), (300, 340); and four expected dual-frequency passband targets: [(240, 250), (300, 310)], [(240, 260), (300, 320)], [(240, 270), (300, 330)], [(240, 280), (300, 340)]. Three types of ablation studies were conducted using a type-2 DFC topology comprising four resonators \cite{r6}. The initial states before optimization were generated using the BRI method.

\subsubsection{Action Space}\label{sec:4-2-1}

\begin{figure}[t!]
    \begin{center}
        \includegraphics[width=\linewidth]{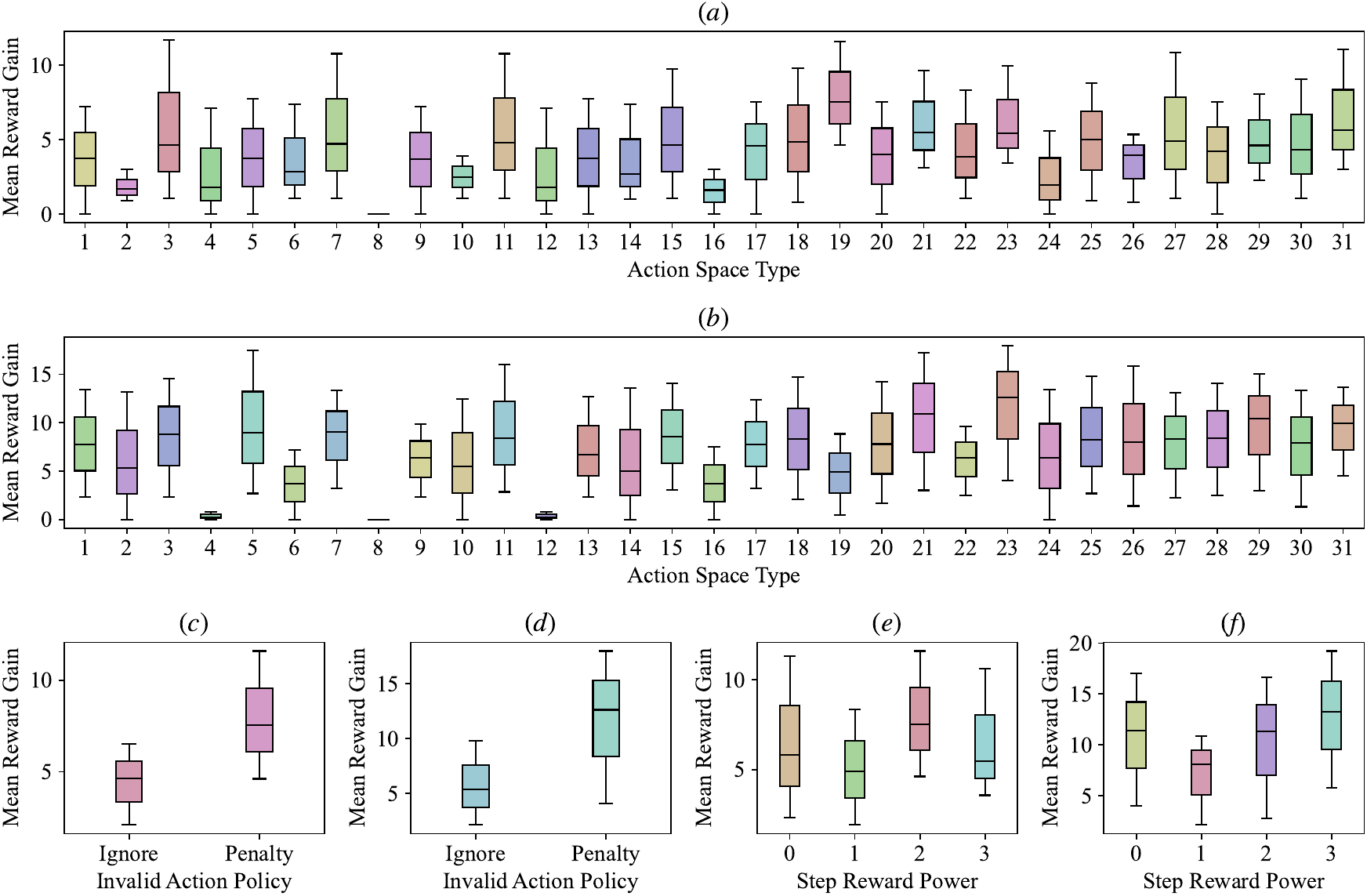}
    \end{center}
    \caption{Ablation study results. The central black lines in the boxes indicate the mean values. Average reward gains for 31 types of selectable action spaces across four \textbf{(a)} single- and \textbf{(b)} dual-frequency passbands. Average reward gains for the two invalid action strategies across the four \textbf{(c)} single- and \textbf{(d)} dual-frequency passbands. Average reward gains for four-step reward power variations across four \textbf{(e)} single- and \textbf{(f)} dual-frequency passbands.}
    \label{fig:3}
\end{figure}

To validate the effectiveness of our chosen action space, we conducted an ablation study involving 31 combinations (listed in Table \ref{tab:1}) comprising parameters $x$, $y$, $a$, $w$, and $u$ of the resonators as potential actions for the agent. The average change in the weighted sum of the IOU and insertion loss were measured before and after optimization for each combination, as depicted in Figs.\ref{fig:3}(a) and (b). This corroborates the discussion in Section \ref{sec:3-2}.

\begin{table}[t!]
\centering
\caption{Different types of selectable action spaces.}
\label{tab:1}
\resizebox{\textwidth}{!}{%
\begin{tabular}{cccll}
\toprule
\#Combination &
State Space &
Action Space &
Operable Parameters &
Description \\
\midrule
1  & (4, 1) & 9  & $x$                      & select $x$  \\
2  & (4, 1) & 9  & $y$                      & select $y$  \\
3  & (4, 2) & 17 & $x$, $y$                 & select $x$, $y$  \\
4  & (4, 1) & 3  & $l$                      & select $l$  \\
5  & (4, 2) & 11 & $x$, $l$                 & select $x$, $l$  \\
6  & (4, 2) & 11 & $y$, $l$                 & select $y$, $l$  \\
7  & (4, 3) & 19 & $x$, $y$, $l$            & select $x$, $y$, $a$  \\
8  & (4, 3) & 3  & $w$                      & select $w$, gap$_h$, gap$_w$  \\
9  & (4, 4) & 11 & $x$, $w$                 & select $x$, $w$, gap$_h$, gap$_w$  \\
10 & (4, 4) & 11 & $y$, $w$                 & select $y$, $w$, gap$_h$, gap$_w$  \\
11 & (4, 5) & 19 & $x$, $y$, $w$            & select $x$, $y$, $w$, gap$_h$, gap$_w$  \\
12 & (4, 4) & 5  & $l$, $w$                 & select $a$, $w$, gap$_h$, gap$_w$  \\
13 & (4, 5) & 13 & $x$, $l$, $w$            & select $x$, $l$, $w$, gap$_h$, gap$_w$  \\
14 & (4, 5) & 13 & $y$, $l$, $w$            & select $y$, $l$, $w$, gap$_h$, gap$_w$  \\
15 & (4, 6) & 21 & $x$, $y$, $l$, $w$       & select $x$, $y$, $l$, $w$, gap$_h$, gap$_w$  \\
16 & (4, 3) & 9  & $u$                      & select $u$, gap$_x$, gap$_y$  \\
17 & (4, 4) & 17 & $x$, $u$                 & select $x$, $u$, gap$_x$, gap$_y$  \\
18 & (4, 4) & 17 & $y$, $u$                 & select $y$, $u$, gap$_x$, gap$_y$  \\
19 & (4, 5) & 25 & $x$, $y$, $u$            & select $x$, $y$, $u$, gap$_x$, gap$_y$  \\
20 & (4, 4) & 11 & $l$, $u$                 & select $a$, $u$, gap$_x$, gap$_y$  \\
21 & (4, 5) & 19 & $x$, $l$, $u$            & select $x$, $l$, $u$, gap$_x$, gap$_y$  \\
22 & (4, 5) & 19 & $y$, $l$, $u$            & select $y$, $l$, $u$, gap$_x$, gap$_y$ \\
23 & (4, 6) & 27 & $x$, $y$, $l$, $u$       & select $x$, $y$, $l$, $u$, gap$_x$, gap$_y$  \\
24 & (4, 6) & 11 & $w$, $u$                 & select $w$, $u$, gap$_h$, gap$_w$, gap$_x$, gap$_y$  \\
25 & (4, 7) & 19 & $x$, $w$, $u$            & select $x$, $w$, $u$, gap$_h$, gap$_w$, gap$_x$, gap$_y$  \\
26 & (4, 7) & 19 & $y$, $w$, $u$            & select $y$, $w$, $u$, gap$_h$, gap$_w$, gap$_x$, gap$_y$  \\
27 & (4, 8) & 27 & $x$, $y$, $w$, $u$       & select $x$, $y$, $w$, $u$, gap$_h$, gap$_w$, gap$_x$, gap$_y$  \\
28 & (4, 7) & 13 & $l$, $w$, $u$            & select $a$, $w$, $u$, gap$_h$, gap$_w$, gap$_x$, gap$_y$  \\
29 & (4, 8) & 21 & $x$, $l$, $w$, $u$       & select $x$, $l$, $w$, $u$, gap$_h$, gap$_w$, gap$_x$, gap$_y$  \\
30 & (4, 8) & 21 & $y$, $l$, $w$, $u$       & select $y$, $l$, $w$, $u$, gap$_h$, gap$_w$, gap$_x$, gap$_y$  \\
31 & (4, 9) & 29 & $x$, $y$, $l$, $w$, $u$  & select $x$, $y$, $l$, $w$, $u$, gap$_h$, gap$_w$, gap$_x$, gap$_y$  \\
\bottomrule
\end{tabular}%
}
\end{table}

\subsubsection{Invalid Action Penalty}\label{sec:4-2-2}

Generally, two strategies are employed to handle invalid actions.
The first strategy ignores invalid actions, ensuring the perpetual legality of layout states. In the short term, this strategy is advantageous because it allows the agent to explore a greater state space. However, it fails to teach the model how to avoid such actions. In the long term, a considerable proportion of attempted actions become invalid.
The second strategy is to penalize invalid actions. When an invalid action is encountered, this strategy terminates the exploration of the agent and resets its environmental state. In the short term, the agent loses its ability to explore additional states. However, in the long term, the agent learns to avoid invalid actions. As the agent continues to learn, more invalid actions are avoided, increasing the number of exploration steps. We conducted an ablation study and compared the results with an invalid action-ignoring strategy to validate the effectiveness of the IDP strategy. The results are shown in Figs.\ref{fig:3}(c) and (d).

\subsubsection{Power of Step Reward}\label{sec:4-2-3}

In the initial stages of implementing the \emph{FilterEnv} environment, we simply used the difference between the rewards before and after a step as the step reward. To encourage the model to learn to make steps that improve the state and avoid steps that worsen it, we conducted an ablation study on the power of the difference between the rewards before and after a step, denoted as $r_1$ and $r_2$, respectively. In particular, if the layout state after the step was illegal, we followed the IDP strategy in Section \ref{sec:3-5-2} to set the step reward to $-200^{power}$. We conducted an ablation study by setting the power to 0, 1, 2, and 3 to avoid the potential risk of reward overflow owing to excessively large rewards. The ablation results are displayed in Figs.\ref{fig:3}(e) and (f).
\begin{equation}\label{eq:3}
  \begin{aligned}
    \mathrm{sign} &= \left\{\begin{array}{ll}
                    1,& \textrm{if $r_1<r_2$}\\
                    -1,& \textrm{otherwise}
                    \end{array}\right.\\
    \mathrm{step}_r &= \mathrm{sign}\times(r_1-r_2)^{power}
  \end{aligned}
\end{equation}

\subsubsection{Hyperparameter Settings}

\begin{table}[t]
\centering
\caption{Experimental results of hyperparameter settings.}
\label{tab:c}
\begin{tabular}{lllll}
\toprule
Hyperparameter settings & $\alpha$=10, $\beta$=166 & $\alpha$=1, $\beta$=10 & $\alpha$=1, $\beta$=9 & $\alpha$=1, $\beta$=11 \\
\midrule
Expected Mean Rewards  & 1143.16 & 129.69 & 124.47 & 119.83 \\
Mean Reward Return Ratio (\%)  & 57.2 & 81.1 & 80.8 & 72.2 \\
\bottomrule
\end{tabular}%
\end{table}

Regarding the selection of hyperparameters, the strategy for selecting $\alpha$ and $\beta$ in Eqs.\ref{eq:2} and \ref{eq:3} is as follows: The role of $\alpha$ is to eliminate the fractional part of $\beta$. We aim for the RL agent to optimize both the passband IOU and insertion loss of the DFC. The effective range for passband IOU is $(0, 100]$, and insertion loss is the absolute value of the maximum value of $s_{21}$, where $s_{21}$'s value range is $(-\infty, 0)$. Typically, $s_{21}$ above -6dB is considered the effective operating bandwidth for a filter circuit, so the range for insertion loss is $(0, 6]$. Initially, we tried setting $\alpha$=10 and $\beta$=166 to equally emphasize the importance of passband IOU and insertion loss. However, we observed that the RL agent overly focused on achieving low insertion loss, resulting in suboptimal overall performance. Therefore, we attempted to reduce the value of $\beta$. Since 10 is preferable, we set $\alpha$=1 and $\beta$=10. This adjustment improved the results significantly. Further, we tried $\alpha$=1 and $\beta$=9, but this led to insertion loss above 3dB, which was also unsatisfactory. When setting $\alpha$=1 and $\beta$=11, the RL agent again tended to focus on optimizing insertion loss, contrary to our overall optimization goals. Ultimately, setting $\alpha$=1 and $\beta$=10 yielded the best results, as detailed in Table \ref{tab:c}.

\subsection{Performance Evaluation}\label{sec:4-3}

\subsubsection{Evaluation on Feedforward Network}\label{sec:4-3-a}

\begin{table}[t!]
\centering
\caption{Comparison results of our feedforward network with CircuitGNN.}
\label{tab:b}
\resizebox{\textwidth}{!}{%
\begin{tabular}{ccccccccc}
\toprule
\multirow{2}{*}{\#Resonators} &
  \multirow{2}{*}{Topology} &
  \multirow{2}{*}{\begin{tabular}[c]{@{}c@{}}\#Samples\\ (train/valid/test)\end{tabular}} &
  \multicolumn{3}{c}{CircuitGNN} &
  \multicolumn{3}{c}{Ours} \\ \cline{4-9}
 &
   &
   &
  Train error (dB) &
  Valid error (dB) &
  Test error (dB) &
  Train error (dB) &
  Valid error (dB) &
  Test error (dB) \\
\midrule
4 & 0   & 12000/1500/1500 & 0.670 & 2.085 & 2.062 & 0.699 & 1.680      & 1.761 \\
4 & 1   & 12000/1500/1500 & 0.685 & 2.325 & 2.329 & 0.786 & 2.064      & 2.017 \\
4 & 2   & 12000/1500/1500 & 0.579 & 1.422 & 1.468 & 0.530 & 1.252      & 1.283 \\
4 & 3   & 12000/1500/1500 & 0.569 & 1.419 & 1.364 & 0.584 & 1.288      & 1.285 \\
4 & 4   & 12000/1500/1500 & 0.534 & 1.294 & 1.279 & 0.593 & 1.197      & 1.182 \\
4 & 5   & 12000/1500/1500 & 0.561 & 1.255 & 1.306 & 0.506 & 1.072      & 1.091 \\
4 & 6   & 12000/1500/1500 & 0.683 & 2.037 & 2.029 & 0.767 & 1.663      & 1.596 \\
4 & 7   & 12000/1500/1500 & 0.628 & 1.592 & 1.597 & 0.585 & 1.445      & 1.43  \\
4 & 8   & 12000/1500/1500 & 0.670 & 1.706 & 1.694 & 0.769 & 1.573      & 1.641 \\
4 & 9   & 12000/1500/1500 & 0.655 & 1.755 & 1.734 & 0.618 & 1.460      & 1.443 \\
4 & avg & -               & 0.623 & 1.689 & 1.686 & 0.644 & 1.469      & 1.473 \\
\hline
5 & 0   & 20000/2500/2500 & 0.713 & 2.044 & 2.101 & 0.684 & 1.637      & 1.710 \\
5 & 1   & 20000/2500/2500 & 0.705 & 1.871 & 1.864 & 0.737 & 1.759      & 1.715 \\
5 & 2   & 20000/2500/2500 & 0.715 & 1.852 & 1.815 & 0.818 & 1.699      & 1.633 \\
5 & 3   & 20000/2500/2500 & 0.710 & 1.857 & 1.823 & 0.743 & 1.531      & 1.591 \\
5 & 4   & 20000/2500/2500 & 0.673 & 1.925 & 1.899 & 0.643 & 1.810      & 1.740 \\
5 & 5   & 20000/2500/2500 & 0.732 & 2.173 & 2.137 & 0.678 & 1.754      & 1.711 \\
5 & 6   & 20000/2500/2500 & 0.769 & 2.510 & 2.525 & 0.902 & 2.058      & 2.029 \\
5 & 7   & 20000/2500/2500 & 0.766 & 2.499 & 2.494 & 0.706 & 2.002      & 2.080 \\
5 & 8   & 20000/2500/2500 & 0.778 & 2.815 & 2.829 & 0.925 & 2.646      & 2.527 \\
5 & avg & -               & 0.729 & 2.172 & 2.165 & 0.766 & 1.877      & 1.860 \\
\hline
3 & 0   & 0/0/1000        & -     & -     & 1.490 & -     & -          & 0.948 \\
3 & 1   & 0/0/1000        & -     & -     & 1.362 & -     & -          & 0.866 \\
3 & 2   & 0/0/1000        & -     & -     & 1.260 & -     & -          & 1.024 \\
3 & 3   & 0/0/1000        & -     & -     & 3.298 & -     & -          & 2.366 \\
3 & avg & -               & -     & -     & 1.853 & -     & -          & 1.301 \\
\hline
6 & 0   & 0/0/900         & -     & -     & 4.073 & -     & -          & 3.844 \\
6 & 1   & 0/0/900         & -     & -     & 3.792 & -     & -          & 2.028 \\
6 & 2   & 0/0/900         & -     & -     & 4.130 & -     & -          & 2.127 \\
6 & 3   & 0/0/900         & -     & -     & 4.103 & -     & -          & 2.662 \\
6 & 4   & 0/0/900         & -     & -     & 2.646 & -     & -          & 1.998 \\
6 & 5   & 0/0/900         & -     & -     & 3.391 & -     & -          & 2.407 \\
6 & avg & -               & -     & -     & 3.689 & -     & -          & 2.511 \\
\bottomrule
\end{tabular}%
}
\end{table}

Table \ref{tab:b} shows the comparison results of our feedforward network with CircuitGNN \cite{r6}. The table indicates that our network achieved low training and validation errors on DFCs composed of 4 and 5 resonators. This demonstrates that our feedforward network can correctly fit the data and capture the electromagnetic characteristics of circuits with 4 or 5 resonators. The results show that our feedforward network outperforms CircuitGNN on DFCs with all topologies.

In terms of generalization, we considered both the DFC topologies that appear in the training set (\ie DFCs composed of 4 or 5 resonators) and those that do not (\ie DFCs composed of 3 or 6 resonators). As shown in Table \ref{tab:b}, the errors are still relatively low even for circuits the feedforward network has never seen before. This indicates the better generalization capability of our feedforward network, providing strong support for discovering new circuit topologies in future work.

\subsubsection{Cumulative Distribution Function (CDF)}\label{sec:4-3-1}

We compared our method with the inverse optimization method proposed in CircuitGNN \cite{r6} and the commercial finite element solver HFSS \cite{r48} by designing 1500 single-frequency bandpass filter circuits. The bandwidths of the filter circuits were randomly distributed between 10 and 80, as presented in Table \ref{tab:2}, whereas the center frequencies varied randomly between 230 + bandwidth/2 GHz and 370 – bandwidth/2 GHz.

\begin{table}[t!]
\centering
\caption{Passband distribution of the filter circuits used in the experiment.}
\label{tab:2}
\begin{tabular}{lllll}
\toprule
Bandwidth & 10-19 GHz & 20-40 GHz & 41-59 GHz & 60-80 GHz \\
\midrule
Number  & 500 & 200 & 300 & 500  \\
\bottomrule
\end{tabular}%
\end{table}

\begin{figure}[t!]
    \begin{center}
        \includegraphics[width=\linewidth]{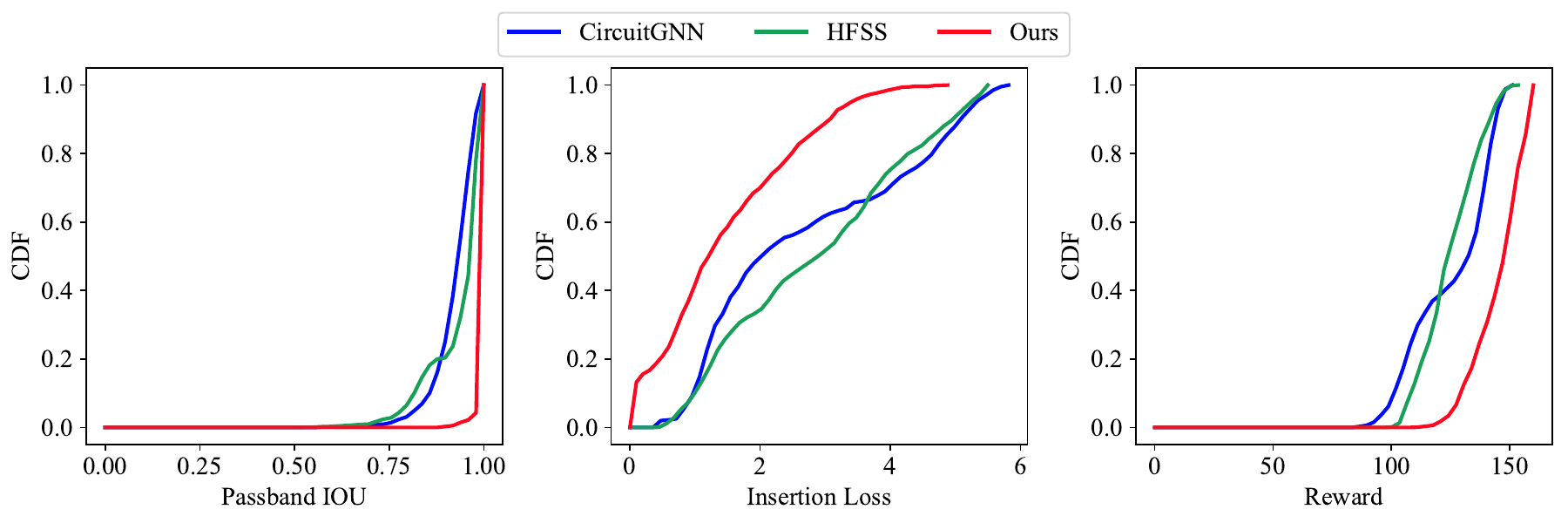}
    \end{center}
    \caption{Comparison of CDF of our method with CircuitGNN and HFSS.}
    \label{fig:4}
\end{figure}

We utilized three metrics to evaluate the quality of the designed filter circuits across 1500 single-frequency bandpass filter circuits. The passband IOU metric assessed the alignment between the delivered passband and the expected passband, following the definition in Section \ref{sec:3-3}. The insertion loss, as defined by \cite{r6}, was calculated as the maximum absolute value of $s_{21}$. Lower insertion loss values indicate superior filter circuit performance. The reward, as defined in Eq.\ref{eq:2}, was introduced to balance the importance of the passband IOU and the insertion loss. This metric provides a more comprehensive representation of the quality of filter circuit design. Note that the circuits generated by the CircuitGNN method often have low insertion losses (good) and low passband IOUs (bad). We opted against the CircuitGNN method, which selects the layout scheme with the lowest train loss; instead, we compared our approach by selecting the layout with the maximum weighted sum of the passband IOU and insertion loss.

The proposed method demonstrates superior circuit performance compared with CircuitGNN and HFSS (Fig.\ref{fig:4}). First, the average passband IOU of the proposed method in 1500 random single-frequency bandpass filter circuit design tasks is 99.5\%, which is significantly higher than the 92.1\% of CircuitGNN and 93.2\% of HFSS. A high passband IOU signifies a good match between the passbands of the delivered filter circuit and the expected design requirements and also implies a steep passband, which enhances the ability of the filter circuit to reject signals from adjacent frequencies. Since our method attains high passband IOU metrics, it can yield filter circuits with superior performance. In addition, the insertion loss of the proposed method is also significantly lower, with an average insertion loss of only 1.44dB, which is much lower than the 2.67dB of CircuitGNN and 2.82dB of HFSS. A lower loss leads to stronger signals at the receiving end, a lower signal-to-noise ratio, and improved reception performance. The insertion loss reduction also enables the transmission end to utilize lower output power. Reducing the transmission power also reduces the power consumption of the system. Finally, considering the combined effect of IOU and loss, our reward function score reached 145.18, whereas CircuitGNN and HFSS scored 125.46 and 124.999, respectively, indicating a significant overall performance enhancement.

\subsubsection{Performance on All Filter Circuit Templates}\label{sec:4-3-2}

\begin{figure}[t!]
    \begin{center}
        \includegraphics[width=\linewidth]{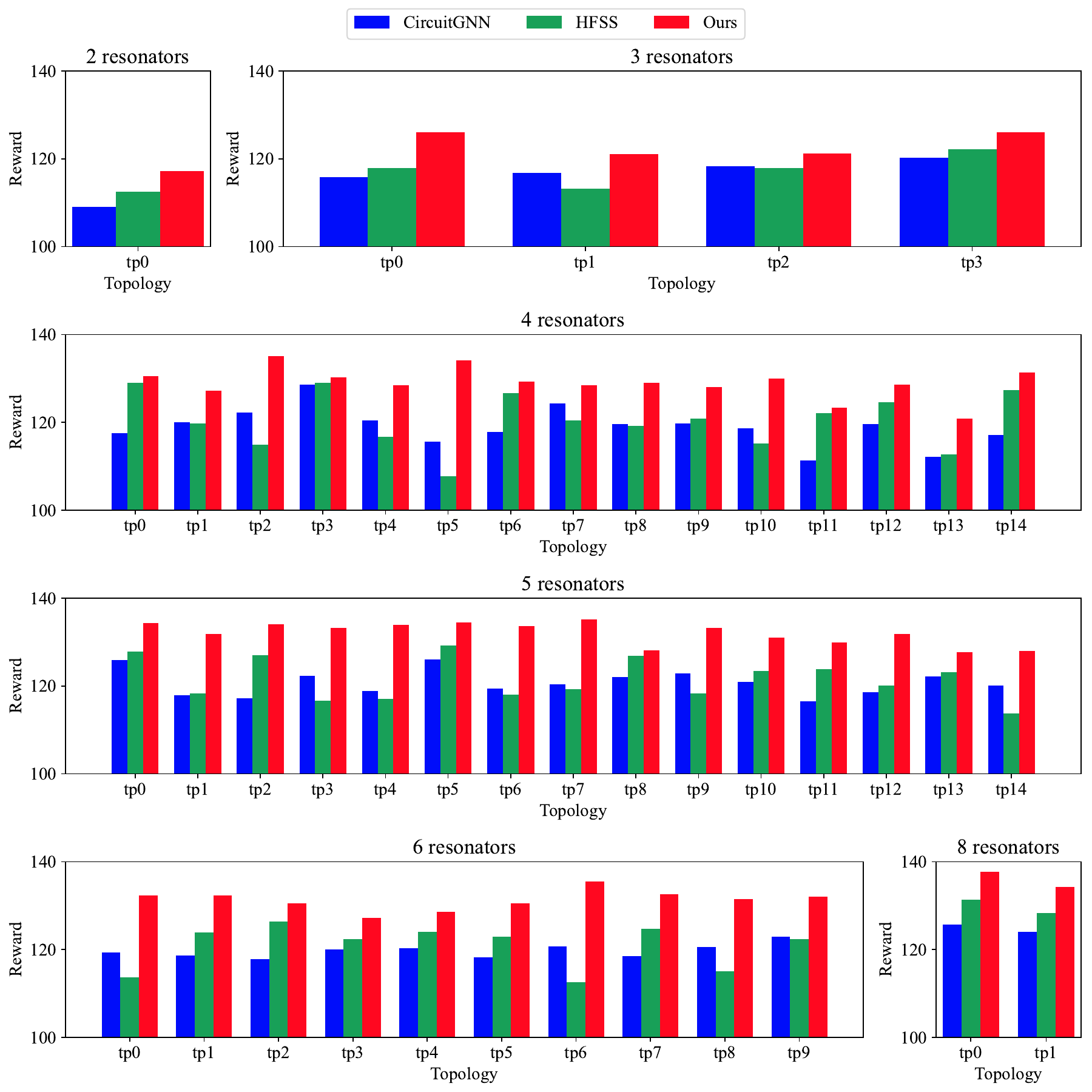}
    \end{center}
    \caption{Comparison of our method with CircuitGNN and HFSS in the design results across all filter circuit templates.}
    \label{fig:5}
\end{figure}

Six single-frequency bandpass filter circuit design tasks were randomly selected, and optimization was performed using our proposed method, HFSS, and the inverse method proposed in CircuitGNN. Each single-frequency bandpass design task was executed across all six filter circuit templates, and the average rewards for the different resonators and topologies were compared. As suggested by the results shown in Fig.\ref{fig:5}, the proposed method is advantageous compared to the CircuitGNN inverse method and the HFSS, consistently exhibiting superior performance and indicating its robustness in filter circuit design tasks. Generally, a single-frequency bandpass filter circuit design requires balancing conflicting objectives while maximizing the passband IOU. The proposed method realizes effective tradeoffs, demonstrating its outstanding optimization capability. Additionally, as the number of resonators increases, the superiority of the proposed method becomes more pronounced. Hence, the proposed method can effectively handle intricate scenarios. Even with the same number of resonators, the proposed method exhibits superior performance across various topologies. For instance, with four resonators, the proposed method significantly outperforms the CircuitGNN inverse method and the HFSS across 15 different topologies that cover diverse coupling schemes and physical layouts between resonators. Thus, the proposed method is generally effective and adaptable to the specifics of each unique filter circuit design template. Electronics engineers can use the proposed method to significantly reduce the time and computational costs required to achieve optimal designs. Consequently, more efficient optimization enables manufacturers to respond swiftly to market demands and technological innovations.

\subsubsection{Performance of Individual Circuit Optimization}\label{sec:4-3-3}

\begin{figure}[t!]
    \begin{center}
        \includegraphics[width=\linewidth]{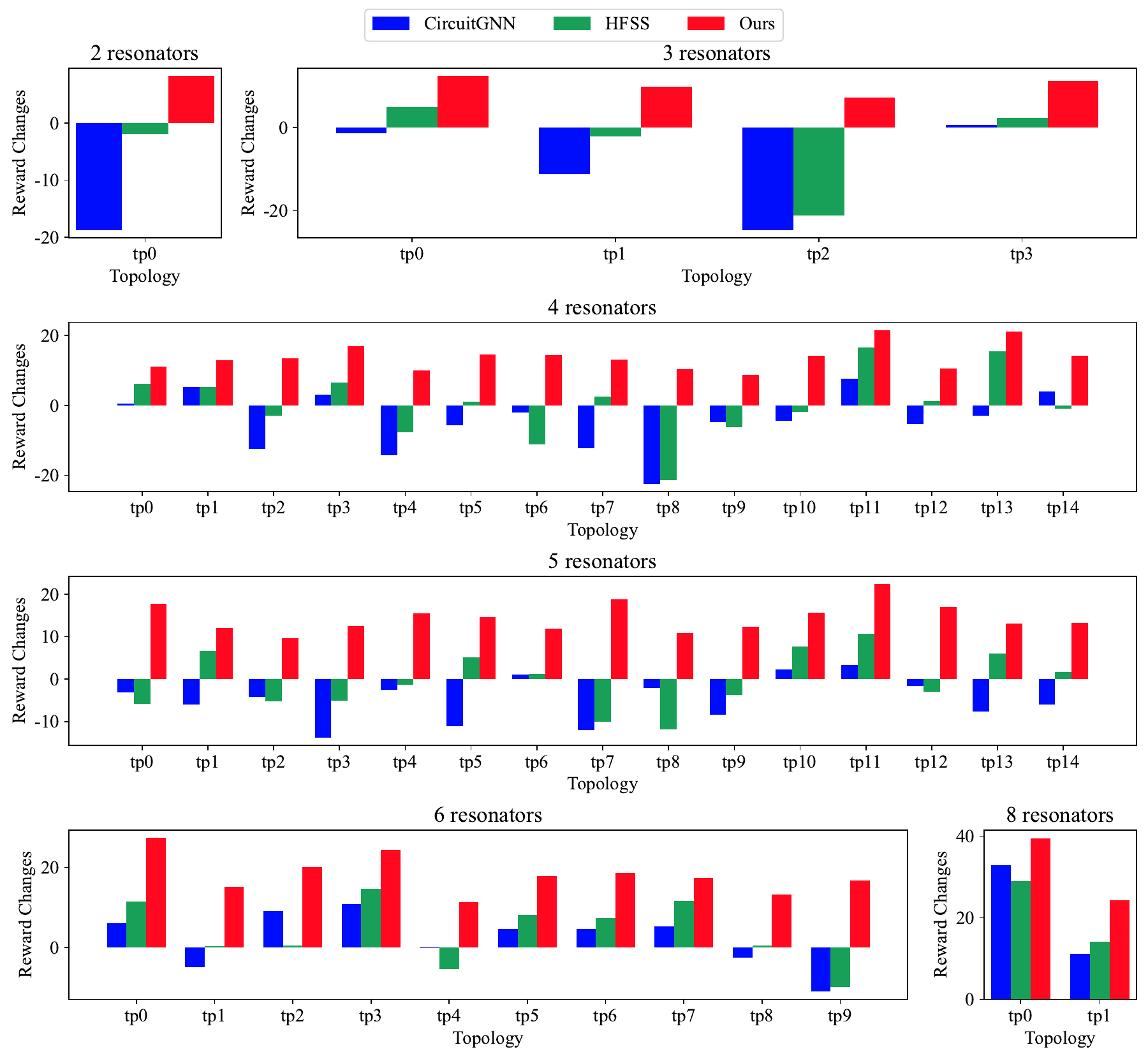}
    \end{center}
    \caption{Comparison of reward changes of our method with CircuitGNN and HFSS before and after optimization across all filter circuit templates.}
    \label{fig:6}
\end{figure}

\begin{figure}[t!]
    \begin{center}
    \includegraphics[width=0.85\linewidth]{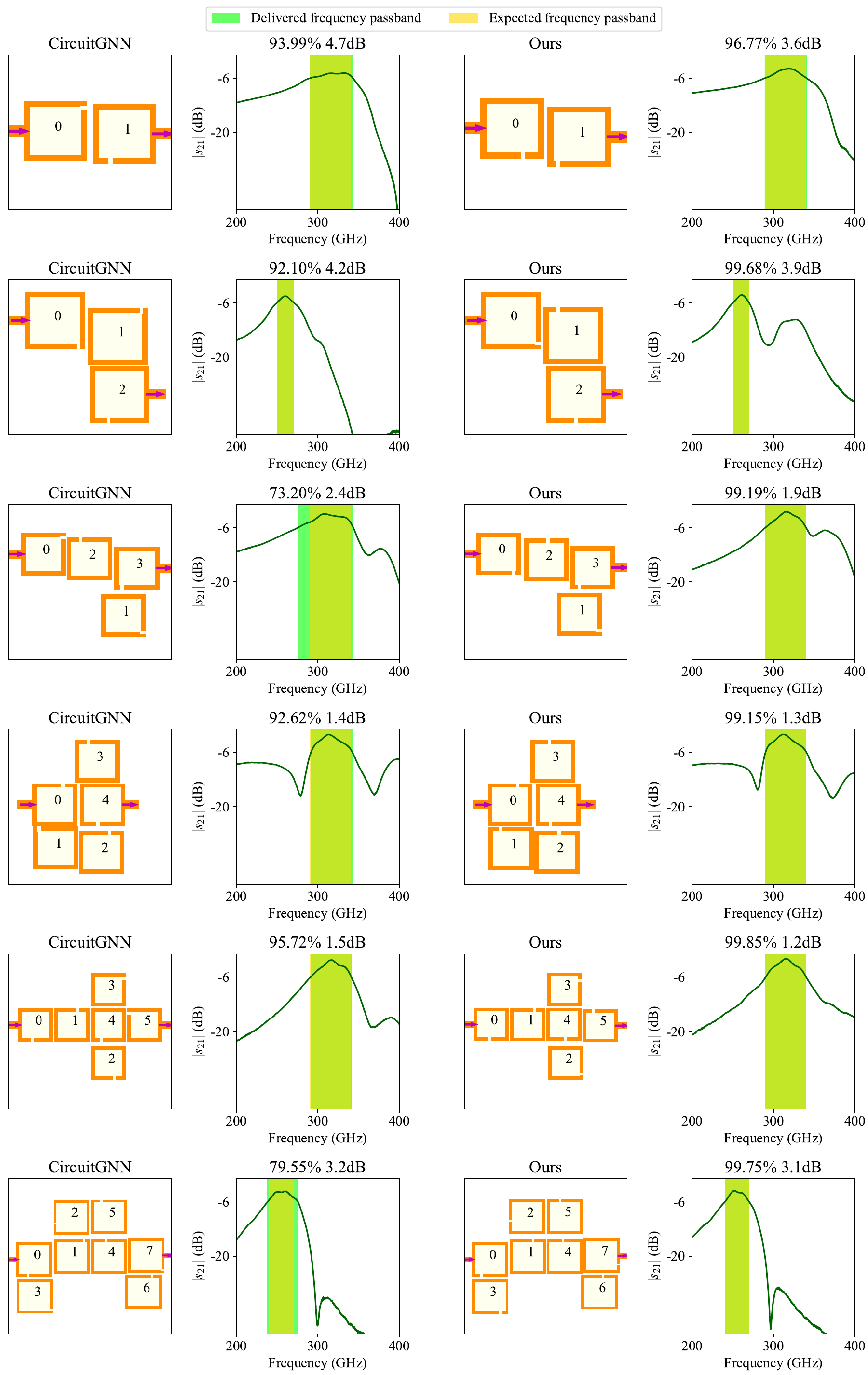}
    \end{center}
    \caption{Comparison of the design results of our method with those of CircuitGNN on different resonator topologies. Left column: Design results of CircuitGNN. Right: Design results of our method. From top to bottom: the first topology (tp0) composed of 2 resonators, the fourth topology (tp3) composed of 3 resonators, the eighth topology (tp7) composed of 4 resonators, the third topology (tp2) composed of 5 resonators, the eighth topology (tp7) composed of 6 resonators, and the second topology (tp1) composed of 8 resonators.}
    \label{fig:8}
\end{figure}

We employed the BRI method to generate the initial layouts for two, three, four, five, six, and eight resonators and the corresponding topologies (\ie all filter circuit templates) for six random single-frequency bandpass filter circuit design tasks. These layouts served as the basis for optimization using the proposed method, HFSS, and the CircuitGNN inverse method. Then, the average reward changes for the different resonators and topologies of the six single-frequency bandpass filter circuit design tasks were compared. In this optimization task, all the compared methods utilized the same initial layout (Fig.\ref{fig:6}). After optimizing using our method, the performance of the filter circuit has been significantly improved, while the HFSS and the inverse optimization method of CircuitGNN have achieved worse results than the initial layout. Fig.\ref{fig:8} shows a comparison of the design results of our method with CircuitGNN and HFSS on topologies composed of different resonators. This fully proves that the stepwise optimization process of our method can indeed continuously improve the design of individual circuits. However, the results of CircuitGNN and HFSS are the opposite. Its passband IOU, interpolation loss, and reward score all degrade during optimization. This may be due to issues with its optimization process, which prevents effective progressive improvement. In contrast, our method simulates an actual design optimization iteration process and can start from an existing design like an electronic engineer and gradually push performance to the limit by adjusting parameters. Adequate training enables the RL agent (\aka the policy network) to gain experience in improving circuit performance. We also use multiple metrics to drive more rewards. Therefore, the proposed method can not only automatically design a new filter circuit but also continue to optimize based on the work of electronic engineers, and the performance is significantly improved.

\subsubsection{Resource Utilization and Optimization Efficiency}\label{sec:4-3-4}

\begin{table}[t]
\centering
\caption{Comparison of the operational efficiencies of our method and CircuitGNN.}
\label{tab:3}
\begin{tabular}{lll}
\toprule
Platform & CircuitGNN & Ours \\
\midrule
CPU (i7-8700T ES)  & $\sim$0.03 iteration/s & $\sim$60 iteration/s  \\
GPU (GTX 3090 24G)  & $\sim$0.83 iteration/s & $\sim$200 iteration/s  \\
\bottomrule
\end{tabular}%
\end{table}

As derived from the results in Section \ref{sec:4-3-3}, the CircuitGNN inverse method \cite{r6} can concurrently optimize 2000 layouts. Thus, this technique demands a high memory with an average occupancy of 2460 MB, whereas the proposed method only requires 866 MB. The proposed method imposes low hardware-performance demands and achieves superior execution efficiency (Table \ref{tab:3}). Low resource utilization implies the potential deployment of our method on cost-effective hardware. By contrast, CircuitGNN requires high-end GPU platforms. Hence, the method proposed in this study significantly reduces machine costs. Furthermore, the proposed approach can optimize individual circuits while enhancing computational efficiency. The training speed reaches approximately 60 iterations per second, which is 1800 times as fast as that of CircuitGNN. Overall, the proposed method can automatically optimize a single circuit efficiently and reliably using ordinary hardware by optimizing the computational process, thereby significantly enhancing the applicability of the proposed method in practical engineering and lowering the threshold for users to benefit from an automated design. The continual optimization of our automated design system could further reduce the demand for computational resources.

\subsection{Discussion}\label{sec:4-4}

\begin{figure}[t!]
    \begin{center}
        \includegraphics[width=\linewidth]{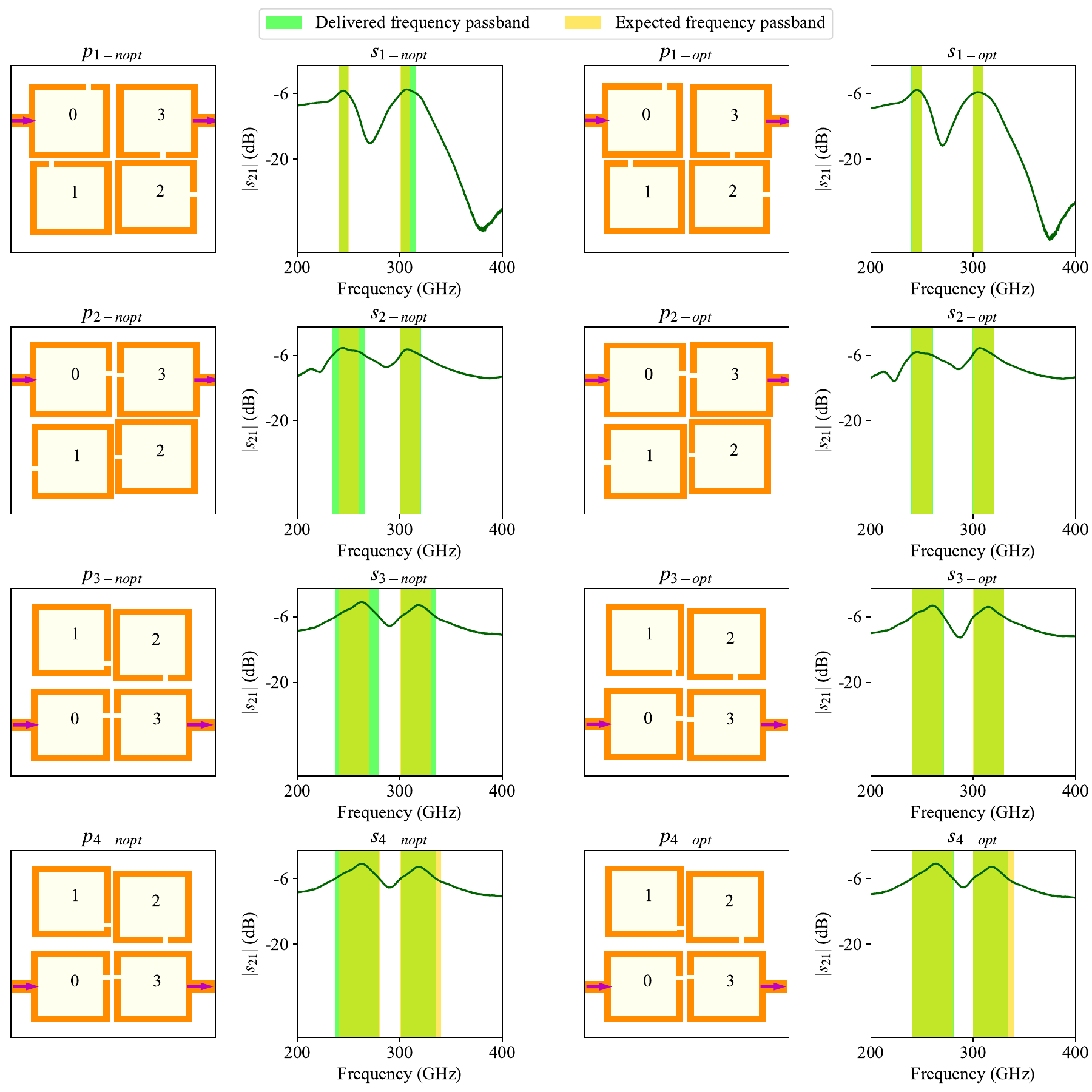}
    \end{center}
    \caption{Dual-passband optimization examples of four instances: layouts before and after optimization, $s_{21}$, and passband. Here, $p_{i-nopt}$ and $s_{i-nopt}$ respectively represent the i-th layout and performance ($s_{21}$ and passband) before optimization, whereas $p_{i-opt}$ and $s_{i-opt}$ represent the layout and performance ($s_{21}$ and passband) after optimization.}
    \label{fig:7}
\end{figure}

\begin{table}[t!]
\centering
\caption{Comparison of passband IOU and insertion loss results before and after optimization for four dual-passband examples using our method.}
\label{tab:4}
\begin{tabular}{lll}
\toprule
Expected passband & Passband IOU & Insertion loss \\
\midrule
$[(240, 250),(300, 310)]$  & 73.22\%/97.68\% & 5.13 dB/5.18 dB  \\
$[(240, 260),(300, 320)]$  & 80.13\%/95.45\% & 4.44 dB/4.46 dB  \\
$[(240, 270),(300, 330)]$  & 76.70\%/97.28\% & 2.81 dB/3.57 dB  \\
$[(240, 280),(300, 340)]$  & 88.28\%/91.75\% & 2.81 dB/2.78 dB  \\
\bottomrule
\end{tabular}%
\end{table}

Moreover, the proposed method demonstrates robust capabilities in simultaneously optimizing multiple passbands. Fig.\ref{fig:7} illustrates the dual-passband cases, and Table \ref{tab:4} presents the specific data. Through optimization, the average passband IOU increased from 79.58\% to 95.54\%, with almost the same average insertion loss. Thus, the proposed method can effectively enhance the overall performance of multi-bandpass filter circuits. The improvement in performance can be attributed to adopting an IOU metric that accurately reflects interactions across multiple passbands. As the number of frequency passbands increases, the complexity of the filter circuit design increases exponentially. The proposed method offers a reliable and automated solution for the design of multi-bandpass filter circuits. Through continual improvements, the proposed method can assist engineers in designing increasingly complex, multifunctional, and multi-frequency filter circuits.

\section{Conclusion}\label{sec:5}

This study demonstrated the feasibility of using RL to optimize high-frequency filter circuits and proposed an RL-based method for the automated optimization of DFCs. We further developed an end-to-end solution for an automated DFC design by integrating BRI, significantly improving the efficiency of the design process and the reliability of the circuit performance. Compared with the existing methods, the proposed approach provides significant improvements. We extended the method to multi-bandpass filter circuit design tasks and achieved outstanding results. The application of RL to circuit design not only enhances design accuracy, but also introduces new design paradigms in electronic engineering. However, several issues remain unresolved. First, better definition and evaluation of the multi-bandpass circuit performance are necessary. Second, the actions that can be performed are limited and may not accommodate non-standardized circuit design requirements. Moreover, we simplified the environmental space and used the geometric parameters of all the resonators comprising the filter circuit as information for interaction with the RL agent. These measures lead to a loss of coupling information between resonators in the real world. Our future work will focus on addressing these limitations to achieve more comprehensive model structures, training methods, and benchmarks. We may employ graph structures to express the resonator features and coupling relationships. Moreover, we will attempt to optimize the rewards designed for multi-passband tasks and continue expanding the action space to optimize the designs.

\bibliographystyle{num}
\bibliography{bibliography}

\end{document}